\documentclass{article}

\usepackage[final, nonatbib]{neurips_2021}
\usepackage[utf8]{inputenc} 
\usepackage[T1]{fontenc}    
\usepackage{hyperref}       
\usepackage{url}            
\usepackage{booktabs}       
\usepackage{amsfonts}       
\usepackage{nicefrac}       
\usepackage{microtype}      
\usepackage{xcolor}         

\usepackage{amsmath}
\usepackage{graphicx}
\usepackage{caption}
\usepackage{subcaption}
\usepackage{amsfonts}
\usepackage{verbatim}
\usepackage[colorinlistoftodos]{todonotes}
\usepackage{authblk}
\usepackage{bm}
\usepackage{bbold}
\usepackage{multirow}
\usepackage{wrapfig}

\usepackage[linesnumbered]{algorithm2e}
\usepackage{algorithmic}

\usepackage{amsthm}

\newtheorem{theorem}{Theorem}

\newtheorem{property}{Property}

\mathchardef\mhyphen="2D 

\newcommand{\RR}{\mathbb{R}}
\newcommand{\EE}{\mathbb{E}}
\newcommand{\BB}{\mathcal{B}}
\newcommand{\DD}{\mathcal{D}}

\newcommand{\PP}{\mathcal{P}}
\newcommand{\VV}{\mathcal{V}}
\newcommand{\HH}{\mathcal{H}}
\newcommand{\LL}{\mathcal{L}}
\newcommand{\TT}{\mathcal{T}}

\newcommand{\argmax}{\mathrm{argmax~}}

\title{Fitting summary statistics of neural data with a differentiable spiking network simulator}

\makeatletter
\let\newtitle\@title
\makeatother

\author{%
  Guillaume Bellec\thanks{Equal contributions. ${}^\diamond$ Senior authors.}~,
  Shuqi Wang${}^*$, Alireza Modirshanechi, Johanni Brea${}^\diamond$, Wulfram Gerstner${}^\diamond$\\
  Laboratory of Computational Neuroscience\\
  École polytechnique fédérale de Lausanne (EPFL) \\
  \texttt{first.lastname@epfl.ch} \\
}

\begin{document}

\maketitle

\begin{abstract}
Fitting network models to neural activity is an important tool in neuroscience. A popular approach is to model a brain area with a probabilistic recurrent spiking network whose parameters maximize the likelihood of the recorded activity. Although this is widely used, we show that the resulting model does not produce realistic neural activity. To correct for this, we suggest to augment the log-likelihood with terms that measure the dissimilarity between simulated and recorded activity. This dissimilarity is defined via summary statistics commonly used in neuroscience and the optimization is efficient because it relies on back-propagation through the stochastically simulated spike trains. We analyze this method theoretically and show empirically that it generates more realistic activity statistics. We find that it improves upon other fitting algorithms for spiking network models like GLMs (Generalized Linear Models) which do not usually rely on back-propagation. This new fitting algorithm also enables the consideration of hidden neurons which is otherwise notoriously hard, and we show that it can be crucial when trying to infer the network connectivity from spike recordings.
\end{abstract}

\section{Introduction}

Modelling neural recordings has been a fundamental tool to advance our understanding of the brain. It is now possible to fit recurrent spiking neural networks (RSNNs) to recorded spiking activity~\cite{pillow2008spatio, runyan2017distinct,rikhye2018thalamic,gerhard2013successful,kobayashi2019reconstructing}.
The resulting network models are used to study neural properties \cite{mensi2012parameter,pozzorini2013temporal, teeter2018generalized,deny2017multiplexed} or to reconstruct the anatomical circuitry of biological neural networks \cite{gerhard2013successful,kobayashi2019reconstructing,das2020systematic}. 

Traditionally a biological RSNN is modelled using a specific Generalized Linear Model~\cite{pillow2008neural} (GLM) often referred to as the Spike Response Model (SRM) \cite{gerstner1995time,pillow2008neural,gerstner2014neuronal}.
The parameters of this RSNN model are fitted to data with the maximum likelihood estimator (MLE).
The MLE is consistent, meaning that if the amount of recorded data becomes infinite it converges to the true network parameters when they exist. However, when fitting neural activity in practice the MLE solutions are often reported to generate unrealistic activity~\cite{mahuas2020new,hocker2017multistep,gerhard2017stability} showing that this method is not perfect despite it's popularity in neuroscience.
We also find that these unrealistic solution emerge more easily when hidden neurons outside of the recorded units have a substantial impact on the recorded neurons. This is particularly problematic because the likelihood is not tractable with hidden neurons and it raises the need for new methods to tackle the problem.


To address this, we optimize \emph{sample-and-measure} loss functions in addition to the likelihood:
these loss functions require \emph{sampling} spiking data from the model and \emph{measuring} the dissimilarity between recorded and simulated data.
To measure this dissimilarity we suggest to compare summary statistics popular in neuroscience like the peristimulus time histogram (PSTH) and the noise-correlation (NC).
Without hidden neurons, this method constrains the network to generate realistic neural activity but without biasing the MLE solution in the theoretical limit of infinite data.
In practice, it leads to network models generating more realistic activity than the MLE.
With hidden neurons, the sample-and-measure loss functions can be approximated efficiently whereas the likelihood function is intractable.
Although recovering the exact network connectivity from the recorded spikes remains difficult \cite{das2020systematic}, we show on artificial data that modelling hidden neurons in this way is crucial to recover the connectivity parameters. In comparison, methods like MLE which ignore the hidden activity wrongly estimate the connectivity matrix.

In practice the method is simple to optimize with automatic differentiation but there were theoretical and technical barriers which have prevented earlier attempts.
The first necessary component is to design an efficient implementation of back-propagation in stochastic RSNN inspired by straight-through gradient estimators \cite{bengio2013estimating,raiko2014techniques} and numerical tricks from deterministic RSNNs \cite{bellec2018long}.
Previous generative models of spiking activity relying on back-propagation used artificial neural networks \cite{sussillo2016lfads,ramesh2019adversarial} which are not interpretable model in terms of connectivity and neural dynamics.
Previous attempts to include hidden neurons in RSNN models did not rely on back-prop but relied on expectation maximization \cite{pillow2008neural,lawhern2010population} or reinforce-style gradients \cite{brea2011sequence,brea2013matching, jimenez2014stochastic, arribas2020rescuing} which have a higher variance \cite{rezende2014stochastic}.
There exist other methods to fit neural data using back-propagation and deep learning frameworks but they do not back-propagate through the RSNN simulator itself, rather they require to engineer and train a separate deep network to estimate a posterior distribution \cite{papamakarios2016fast, lueckmann2017flexible,gonccalves2020training,bittner2019interrogating} or as a GAN discriminator \cite{goodfellow2014generative,molano2018synthesizing,ramesh2019adversarial}.
The absence of a discriminator in the sample-and-measure loss function connects it with other simple generative techniques used outside of the context of neural data \cite{li2015generative,engel2020ddsp,cranmer2020frontier}.

Our implementation of the algorithm is published online openly \footnote{Code repository: {\tt \url{https://github.com/EPFL-LCN/pub-bellec-wang-2021-sample-and-measure}}}.

\section{A recurrent spiking neural network (RSNN) model}

We will compare different fitting techniques using datasets of spiking neural activity. 
We denote a tensor of $K^{\mathcal{D}}$ recorded spike trains as $\bm z^{\mathcal{D}} \in \{0, 1\}^{K^{\mathcal{D}} \times T \times n_\VV}$ where $n_\VV$ is the total number of visible neurons recorded simultaneously and $T$ is the number of time steps.
To model the biological network which produced that activity, we consider a simple model that can capture the recurrent interactions between neurons and the intrinsic dynamics of each neuron. 
This recurrent network contains $n_{\VV + \HH}$ neurons connected arbitrarily and split into a visible and a hidden population of sizes $n_\VV$ and $n_\HH$.
Similarly to \cite{pillow2008spatio,gerhard2013successful,macke2011empirical,mahuas2020new} we use a GLM where each unit is modelled with a SRM neuron \cite{gerstner1995time} with $u_{t,j}$ being the distance to the threshold of neuron $j$ and its spike $z_{t,j}$ is sampled at time step $t$ from a Bernoulli distribution $\mathcal{B}$ with mean $\sigma(u_{t,j})$ where $\sigma$ is the sigmoid function. The dynamics of the stochastic recurrent spiking neural network (RSNN) are described by:
\begin{eqnarray}
 z_{t,j} & \sim & \mathcal{B} \left(\sigma (u_{t,j})\right) \text{\hspace{1cm} with \hspace{1cm}} u_{t,j} = \frac{v_{t,j} - v_\mathrm{thr}}{v_\mathrm{thr}}\label{eq:1} \\
 v_{t,j} & = & \sum_{i=1}^{n_{\VV +  \HH}} \sum_{d=1}^{d_{\max}} W_{j,i}^d z_{t-d,i} + b_j + \mathcal{C}_{t,j}~,
 \label{eq:2}
\end{eqnarray}
where $\bm W$ defines the spike-history and coupling filters spanning $d_{\max}$ time-bins, $\bm b$ defines the biases, $v_{\mathrm{thr}}=0.4$ is a constant, and $\mathcal{C}$ is a spatio-temporal stimulus filter processing a few movie frames and implemented here as a convolutional neural network (CNN) (this improves the fit accuracy as seen in \cite{mahuas2020new, mcintosh2016deep} and in Figure \ref{fig:cnn} from the appendix).
Equations \eqref{eq:1} and \eqref{eq:2} define the probability $\mathcal{P}_\phi( \bm z)$ of simulating the spike trains $\bm z$ with this model and $\phi$ represents the concatenation of all the network parameters ($\bm W$, $\bm b$ and the CNN parameters). Traditionally the parameters which best explain the data are given by the MLE: $\argmax_\phi \PP_\phi (\bm z^\DD)$ \cite{pillow2008spatio,pozzorini2013temporal,macke2011empirical,kobayashi2019reconstructing,gerhard2013successful}.
When all neurons are visible, the likelihood factorizes as $\prod_{t} \mathcal{P}_\phi( \bm z^{\mathcal{D}}_t| \bm z_{1:t-1}^{\mathcal{D}})$, therefore the log-likelihood can be written as the negative cross-entropy ($CE$) between $\bm z_t^{\mathcal{D}}$ and  $\sigma(\bm u_{t}^\DD) = \PP_\phi(\bm z^{\DD}_t = \bm 1 | \bm z_{1:t-1}^{\mathcal{D}})$ where $\bm u_{t}^\DD$ is computed as in equation \eqref{eq:2} with $\bm z = \bm z^\DD$.
So when all neurons are recorded and $\bm z^\DD$ are provided in the dataset the computation of the MLE never needs to simulate spikes from the model and it is sufficient to minimize the loss function:
\begin{equation}
 \mathcal{L}_{MLE} = - \log \PP_\phi (\bm z^\DD) = CE(\bm z^\DD, \sigma(\bm u ^\DD))~.
 \label{eq:3}
\end{equation}

%

\section{The sample-and-measure loss functions}

\begin{figure}
\includegraphics[width=\textwidth,trim=2 2 2 2,clip]{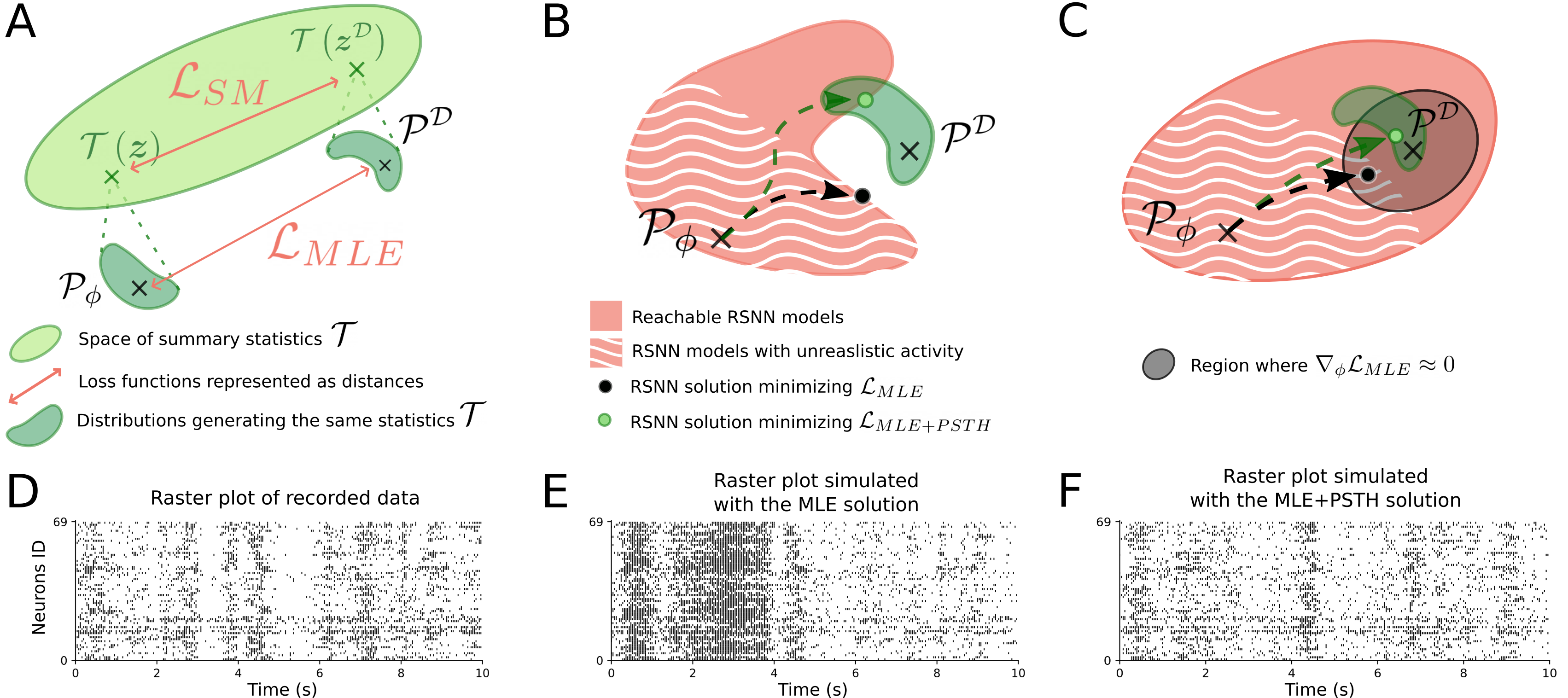}
\caption{
\textbf{A)}
The distribution $\PP_\phi$ represents the RSNN model and $\PP_\DD$ represents the true biological network, the goal is to bring $\PP_\phi$ close to $\PP_\DD$.
The loss function 
$\LL_{MLE}$ is represented as the distance between $\PP_\phi$ and $\PP_{\DD}$ because it is equal up to a constant to $D_{KL}(\PP_{\DD},\PP_{\phi})$.
We draw the space of summary statistics $\TT$ to represent the loss function $\LL_{PSTH}$ as the distance between the statistics $\TT(\bm z)$ simulated from the RSNN model $\PP_\phi$ and measured from the data $\PP_{\DD}$.
\textbf{B)} Even if the model is misspecified and $\PP_\phi$ cannot perfectly match the true distribution $\PP_\DD$, when  $\LL_{MLE+PSTH}$ is minimized the statistics $\TT$ are indistinguishable between simulation and data (the green dot lands in the dark green area).
When minimizing MLE alone, the solution (black dot) might generate unrealistic activity. \textbf{C)} When the RSNN model is expressive enough to represent the true distribution but the data is insufficient, so there is some uncertainty about the true network parameters (black ellipse): minimizing $\LL_{MLE+PSTH}$ favours solutions with realistic PSTH. \textbf{D)} Spikes recorded simultaneously from the V1-dataset \cite{smith2008spatial}. \textbf{E)} Simulated spiking response from an RSNN minimizing $\LL_{MLE}$. \textbf{F)} Same as \textbf{E} but with $\LL_{MLE+PSTH}$.}
\label{fig:geometry}
\end{figure}


In this section we describe the sample-and-measure loss functions which include simulated data in a differentiable fashion in the optimization objective, a direct benefit is to enable the consideration of hidden neurons.
We define the \emph{sample-and-measure} loss functions as those which require \emph{sampling} spike trains $\bm z \in \{0,1\}^{K \times T \times n_{\VV + \HH}}$ from the model $\mathcal{P}_\phi$ and \emph{measuring} the dissimilarity between the recorded and simulated data.
This dissimilarity is defined using some statistics $\TT(\bm z)$ and the generic form of the sample-and-measure loss functions is:
\begin{equation}
\LL_{SM} =
d\Big( ~ \TT( \bm z^\DD), 
~ \EE_{\PP_\phi} \left[ ~\TT( \bm z) ~\right] ~ \Big)~,
\label{eq:sm-with-stats}
\end{equation}
where $d$ is a dissimilarity function, like the mean-squared error or the cross entropy.
To compute the expectations $\EE_{\PP_\phi}$ we use Monte-Carlo estimates from a batch of simulated trials $\bm z$.
For example to match the PSTH between the simulated and recorded data,
we consider the statistics $\TT(\bm z)_{t,i} = \frac{1}{K} \sum_k \bm z_{t,i}^k$ and evaluate the expectation with the unbiased estimate $\bar{\sigma}_{t,i} = \frac{1}{K} \sum_{k} \sigma(u_{i,t}^k)$.
Denoting the PSTH of the data as $\bar{z}_{t,i}^\DD = \frac{1}{K} \sum_k \bm z^{k,\DD}_{t,i}$ and choosing $d$ to be the cross-entropy, we define the sample-and-measure loss function for the PSTH:
\begin{equation}
    \mathcal{L}_{\mathrm{PSTH}} =
    CE({\bar{\bm z}}^\DD,\bar{{\bm \sigma}})~.
    \label{eq:obj-psth}
\end{equation}
When all neurons are visible, we minimize the loss function $\LL_{MLE+SM} = \mu_{MLE} \LL_{MLE} + \mu_{SM} \LL_{SM}$ where $\mu_{MLE},\mu_{SM} >0$ are constant scalars.
When there are hidden neurons, the log-likelihood is intractable.
Instead we minimize the negative of a lower bound of the log-likelihood (see appendix \ref{sec:suppl-sample-and-measure} for a derivation inspired by \cite{jordan1999introduction,brea2011sequence, brea2013matching, jimenez2014stochastic}):
\begin{equation}
    \LL_{ELBO} = CE(\bm z^\DD, \sigma(\bm u^\VV))~,
    \label{eq:elbo}
\end{equation}
with $\sigma(\bm u^\VV)$ being the firing probability of the visible neurons, where the visible spikes $\bm z^\VV$ are clamped to the recorded data $\bm z^\DD$ and the hidden spikes $\bm z^\HH$ are sampled according to the model dynamics.
Hence the implementation of $\LL_{ELBO}$ and $\LL_{SM}$ are very similar with the difference that the samples used in $\LL_{SM}$ are not clamped (but all our results about $\LL_{SM}$ are also valid when they are clamped if we use the extended definition given in Appendix \ref{sec:suppl-sample-and-measure}).
To compute the gradients with respect to these loss functions we use back-propagation which requires the propagation of gradients through the stochastic samples $\bm z$.
If they were continuous random variables, one could use the reparametrization trick~\cite{rezende2014stochastic}, but extending this to discrete distributions is harder \cite{bengio2013estimating,raiko2014techniques, maddison2016concrete,tucker2017rebar,neftci2019surrogate}. One way to approximate these gradients is to relax the discrete dynamics into continuous ones~\cite{maddison2016concrete} or to use relaxed control variates~\cite{tucker2017rebar}, but we expect that the relaxed approximations become more distant from the true spiking dynamics as the network architecture gets very deep or if the network is recurrent.
Instead, we choose to simulate the exact spiking activity in the forward pass and use straight-through gradient estimates \cite{bengio2013estimating,raiko2014techniques} in the backward pass by defining a pseudo-derivative $\frac{\partial z_{t,i}}{\partial u_{t,i}}$ over the binary random variables $z_{t,i} \sim \BB(\sigma(\bm u_{t,i}))$.
We use here the same pseudo-derivative $\frac{\partial z_{t,i}}{\partial u_{t,i}}=\gamma \max (0,1-|u_{t,i}|)$ as in deterministic RNNs~\cite{bellec2018long} because the dampening factor (here $\gamma=0.3$) can avoid the explosive accumulation of approximation errors through the recurrent dynamics~\cite{bengio1994learning}.
Although the resulting gradients are biased, they work well in practice.

\paragraph{A geometrical description of a sample-and-measure loss function}
In the remaining paragraphs of this section we provide a geometrical representation and a mathematical analysis of the loss function $\LL_{SM}$.
For this purpose, we consider that the recorded spike trains $\bm z^\DD$ are sampled from an unknown distribution $\PP_\DD$ and we formalize that our goal is to bring the distribution $\PP_\phi$ as close as possible to $\PP_\DD$.
In this view, we re-write $\LL_{SM}=d(\EE_{\PP_\DD}\left[ \TT(\bm z)\right], \EE_{\PP_\phi}\left[ \TT(\bm z)\right])$ and we re-interpret $\LL_{MLE}$ as the Kullback-Leibler divergence $(D_{KL})$ from $\PP_\phi$ to $\PP_\DD$.
This is equivalent because the divergence $D_{KL}(\PP_\DD, \PP_\phi)$ is equal to $- \EE_{\PP_\DD}\left[ \log \PP_\phi(\bm z^\DD) \right]$ up to a constant.

In Figure \ref{fig:geometry} we represent the losses $\LL_{MLE}$ and $\LL_{PSTH}$ in the space of distributions and we can represent $\LL_{MLE}= D_{KL}(\PP_{\DD},\PP_{\phi})$ as the distance between $\PP_{\DD}$ and $\PP_{\phi}$.
To represent the sample-and-measure loss function $\LL_{SM}$ (or specifically $\LL_{PSTH}$ in Figure \ref{fig:geometry}), we project the two distributions onto the space of summary statistics $\TT$ represented in light green. Hence, these projections represent the expected statistics $\EE_{\PP_\phi} \left[ \TT(\bm z) \right]$ and $\EE_{\PP_\DD} \left[ \TT(\bm z^\DD) \right]$ and $\LL_{SM}$ can be represented as the distance between the two projected statistics.

Although minimizing $\LL_{MLE}$ should recover the true distributions (i.e. the biological network) if the recorded data is sufficient and the model is well specified, these ideal conditions do not seem to apply in practice because the MLE solution often generates unrealistic activity (see Figure \ref{fig:geometry}D-E).
Panels B and C in Figure \ref{fig:geometry} use the geometrical representation of panel A to summarize the two main scenarios where minimizing $\LL_{MLE+SM}$ is better than $\LL_{MLE}$ alone.
In panel B, we describe a first scenario in which the model is misspecified meaning that it is not possible to find $\phi^*$ so that $\mathcal{P}_\DD=\PP_{\phi^*}$.
In this misspecified setting, there is no guarantee that the MLE solution yields truthful activity statistics and it can explain why the MLE solution generates unrealistic activity (Figure \ref{fig:geometry}E).
In this case, adding a sample-and-measure loss function can penalize unrealistic solutions to solve this problem (Figure \ref{fig:geometry}B and F).
Another possible scenario is sketched in panel C. It describes the case where the model is well specified but $\LL_{MLE}$ is flat around $\PP_\DD$ for instance because too few trials are recorded or some neurons are not recorded at all.
In that case we suggest to minimize $\LL_{MLE+SM}$ to nudge the solution towards another optimum where $\LL_{MLE}$ is similarly low but the statistics $\TT$ match precisely. In this sense, $\LL_{SM}$ can act similarly as a Bayesian log-prior to prefer solutions producing truthful activity statistics.

\paragraph{Theoretical analysis of the sample-and-measure loss function}
To describe formal properties of the sample-and-measure loss function $\LL_{SM}$, we say that two distributions are indistinguishable according to the statistics $\TT$ if the expectation $\EE \left[ \TT(\bm z) \right]$ is the same for both distributions. We assume that the dissimilarity function $d(\TT,\TT')$ reaches a minimum if and only if $\TT=\TT'$ (this is true for the mean-squared error and the cross-entropy).
Then for any statistics $\TT$ and associated dissimilarity function $d$ we have:

\begin{property}
If the RSNN model is expressive enough
so that there exists parameters $\phi^\circ$ for which $\PP_\phi$ and $\PP^\DD$ are indistinguishable according to the statistics $\TT$, then $\phi^\circ$ is a global minimum of $\LL_{SM}$. Reciprocally, if this minimum is reached then $\PP_\phi$ and $\PP^\DD$ are indistinguishable according to $\TT$.
\label{prop:1}
\end{property}
\setlength{\intextsep}{0pt}%
\setlength{\columnsep}{5pt}%
\begin{wrapfigure}{r}{0.40\textwidth}
  \begin{center}
    \raisebox{0pt}[\dimexpr\height-0.6\baselineskip\relax]{\includegraphics[width=0.35\textwidth]{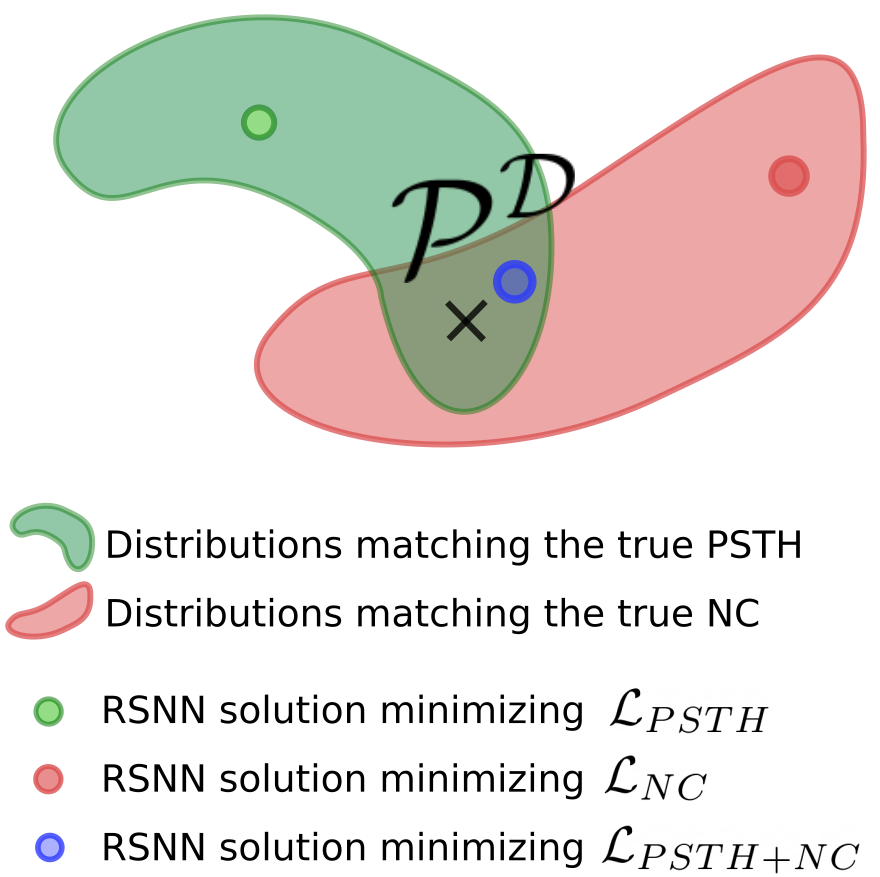}}
  \end{center}
  \caption{Combining $\LL_{PSTH}$ and $\LL_{NC}$.}
  \label{fig:combining-sm}
\end{wrapfigure}
This property is a direct consequence of our assumption on the function $d$. If $\TT$ measures the PSTH it means that the optimized simulator produces the same PSTH as measured in the data.
This can be true even if the model is misspecified which is why we represented in Figure \ref{fig:geometry}B that the RSNN minimizing $\LL_{MLE+PSTH}$ lands in the dark green region where the PSTH of the data is matched accurately.
As one may also want to target other statistics like the noise correlation (NC), it is tempting to consider different statistics $\TT_1$ and $\TT_2$ with corresponding dissimilarity functions $d_1$ and $d_2$ and to minimize the sum of the two losses $\mathcal{L}_{SM_1 + SM_2} = \mu_1 \mathcal{L}_{SM_1} + \mu_2 \mathcal{L}_{SM_2}$ where $\mu_1, \mu_2 > 0$ are constant scalars. Indeed if $d_1$ and $d_2$ follow the same assumption as previously, we have (see Figure \ref{fig:combining-sm} for an illustration):
\begin{property}
If the RSNN model is expressive enough so that there exists $\phi^\circ$ for which $\PP_\DD$ and $\PP_\phi$ are indistinguishable according to both statistics $\TT_1$ and $\TT_2$, then $\phi^\circ$ is a global minimum for $\mathcal{L}_{SM_1 + SM_2}$. Reciprocally, if this minimum is reached, $\PP_\DD$ and $\PP_\phi$ are indistinguishable according to $\TT_1$ and $\TT_2$.
\label{prop:2}
\end{property}
This is again a direct consequence of the assumptions on $d_1$ and $d_2$.
Additionally Figure \ref{fig:geometry}C conveys the idea that $\LL_{SM}$ and $\LL_{MLE}$ are complementary and $\LL_{SM}$ can be interpreted as a log-prior.
This interpretation is justified by the following Property which is similar to an essential Property of Bayesian log-priors. It shows that when it is guaranteed to recover the true model by minimizing $\LL_{MLE}$, minimizing the regularized likelihood $\LL_{MLE+SM}$ will also recover the true model.
\begin{property}
If the RSNN is well specified and identifiable so that $\PP_{\DD}=\PP_{\phi^*}$ and in the limit of infinite data, then the global minimum of $\mathcal{L}_{MLE+SM}$ exists, it is unique and equal to $\phi^*$.
\label{prop:3}
\end{property}
To prove this, we first note that all the conditions are met for the consistency of MLE so $\phi^*$ is the unique global minimum of $\LL_{MLE}$. Also the assumption $\PP_{\DD}=\PP_{\phi^*}$ is stronger than the assumption required in Properties \ref{prop:1} and \ref{prop:2} (previously $\phi^\circ$ only needed to match summary statistics: with parameters $\phi^*$ the model is indeed matching any statistics) so it is also a global minimum of $\LL_{SM}$. As a consequence it provides a global minimum for the summed loss $\LL_{MLE+SM}$. This solution is also unique because it has to minimize $\LL_{MLE}$ which has a unique global minimum. 

It may seem that those properties are weaker than the classical properties of GLMs: in particular $\LL_{SM+MLE}$ is not convex anymore because of $\LL_{SM}$ and the optimization process is not guaranteed to find the global minimum.
This could be a disadvantage for $\LL_{MLE+SM}$ but we never seemed to encounter this issue in practice. In fact, as we argue later when analyzing our simulation results, the optimum of $\LL_{MLE+SM}$ found empirically always seem to be closer to the biological network than the global minimum of $\LL_{MLE}$. We think that it happens because the conditions for the consistency of the MLE and Property 3 (identifiability and infinite data) are not fully met in practice. On the contrary, the Properties 1 and 2 hold with a misspecified model or a limited amount of recorded data which may explain the success of the sample-and-measure loss functions in practice.  See Figure \ref{fig:geometry} for a geometrical interpretation. 
 
\section{Numerical simulations without hidden neurons}
\begin{figure}
  \centering
  \includegraphics[width=\textwidth]{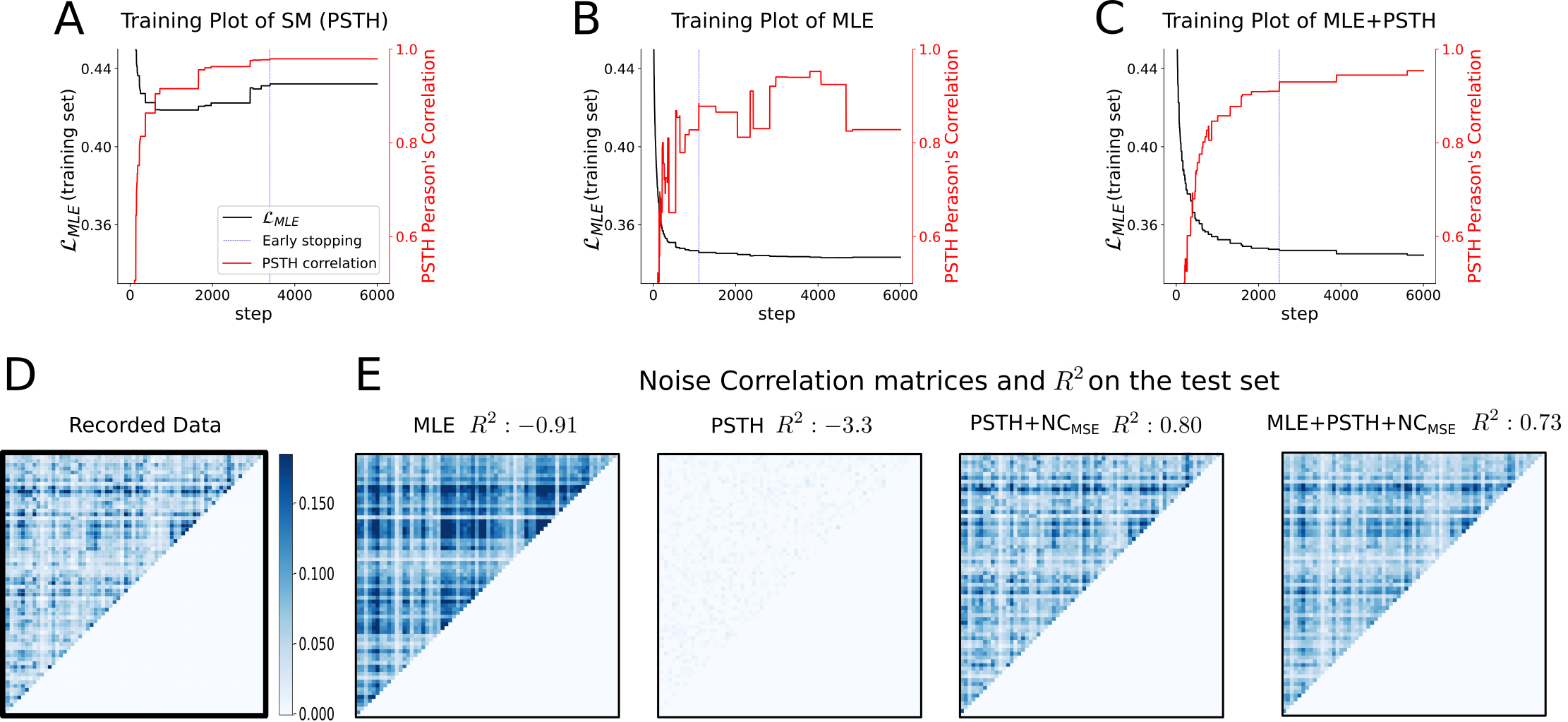}
  \caption{ Learning curves and NC performance summary on the V1-dataset. \textbf{A-C)} Negative log-likelihood (i.e. $\LL_{MLE}$) and PSTH correlation on the training set for three networks trained with $\LL_{PSTH}$, $\LL_{MLE}$ and $\LL_{MLE+PSTH}$.
  The red curves represent the correlation between the PSTH of the recorded and simulated data. To show how the training loss influences the resulting PSTH correlation we plot a new point of the red curve only when the training loss reaches a new minimum. The vertical blue line represents the best network achieving the lowest validation losses (i.e. $\LL_{PSTH}$, $\LL_{MLE}$ and $\LL_{MLE+PSTH}$ for the three corresponding plots).
  \textbf{D}) Noise-correlation (NC) matrix as recorded in the data. x- and y-axis represent the neuron identities. \textbf{E}) NC matrices when the spikes are simulated from the model, it uses the same colorbar as in \textbf{D}). $R^2$ values are reported to compare the recorded NC and the simulated ones, more results are provided in Table~\ref{tab:results_V1}.}
  \label{fig:basic-sm}
\end{figure}

For our first quantitative results we consider a single session of in-vivo recordings from the primary visual cortex of a monkey watching repetitions of the same natural movie~\cite{smith2008spatial}. We refer to this dataset as the V1-dataset. It contains the spike trains of $69$ simultaneously recorded neurons for $120$ repetitions lasting $30$ seconds. We only keep the first $80$ repetitions in our training set and $10$ repetitions are used for early-stopping. Performances are tested on the remaining $30$.
In our first numerical results we do not include hidden neurons.

To illustrate that minimizing $\LL_{MLE}$ alone does not fit well the statistics of interest, we show in Figure \ref{fig:basic-sm}A-C the learning curves obtained when minimizing $\LL_{MLE}$, $\LL_{PSTH}$ and $\LL_{MLE+PSTH}$.
We evaluate the PSTH correlation between simulated and recorded activity every time the training loss function reaches a new minimum. With MLE in Figure \ref{fig:basic-sm}B, the PSTH correlation saturates at a sub-optimal level and drops unexpectedly when $\LL_{MLE}$ decreases.
In contrast, with the sample-and-measure loss function, the PSTH correlation improves monotonously and steadily (see Figure \ref{fig:basic-sm}A).
In Figure \ref{fig:basic-sm}C, one sees that minimizing $\LL_{MLE+SM}$ produces low values of $\LL_{MLE}$ and maximizes efficiently the PSTH correlation.

We then fit simultaneously the PSTH and the noise-correlation (NC) on the V1-dataset.
The NC matrix is complementary to the PSTH and it is used regularly to measure the fit performance \cite{ramesh2019adversarial,macke2011empirical,mahuas2020new}. Its entries can be viewed as a measure of functional connectivity, and each coefficient is defined for the neuron pair $i$ and $j$ as the correlation of their activity.
Concretely it is proportional to the statistics $\TT (\bm z)_{i,j}= \frac{1}{K T} \sum_{k,t} (z_{t,i}^k - \bar{z}_{t,i}) (z_{t,j}^k - \bar{z}_{t,j})$ where $\bar{z}_{t,i}$ is the PSTH (see appendix \ref{sec:app-performance-metrics} for details).
Therefore the natural sample-and-measure loss function for NC is the mean-squared error between the coefficients $\TT (\bm z^\DD)_{i,j}$ and the Monte-carlo estimates
$\frac{1}{K T} \sum_{k,t} (\sigma(u_{t,i}^k) - \bar{\sigma}_{t,i}) (\sigma(u_{t,j}^k) - \bar{\sigma}_{t,j})$.
We denote the resulting loss function as $\LL_{NC_{MSE}}$.
We also tested an alternative loss $\LL_{NC}$ which uses the cross entropy instead of mean-squared error and compares: $\TT(\bm z^\DD)_{i,j}= \frac{1}{K T} \sum_{k,t} z_{t,i}^{k,\DD} z_{t,j}^{k,\DD}$ with the Monte-Carlo estimate $\frac{1}{KT} \sum_{k,t} \sigma (u^k_{t,i}) \sigma (u^k_{t,j})$. 

We compare quantitatively the effects of the loss functions $\LL_{MLE}$, $\LL_{PSTH}$, $\LL_{NC}$ and $\LL_{NC_{MSE}}$ and their combinations on the V1-dataset.
The results are summarized in Table \ref{tab:results_V1} and NC matrices are shown in Figure \ref{fig:basic-sm}E.
The network fitted solely with $\mathcal{L}_{PSTH}$ shows the highest PSTH correlation while its noise correlation is almost zero everywhere (see Figure \ref{fig:basic-sm}E), but this is corrected when adding $\LL_{NC}$ or $\LL_{NC_{MSE}}$.
In fact a network minimizing $\LL_{MLE}$ alone yields lower performance than minimizing $\LL_{PSTH+NC}$ for both metrics.
When combining all losses into $\LL_{MLE+PSTH+NC}$ or $\LL_{MLE+PSTH+NC_{MSE}}$ the log-likelihood on the test set is not compromised and it fits better the PSTH and the NC: the coefficient of determination $R^2$ of the NC matrix improves by a large margin in comparison with the MLE solution.
Analyzing the failure of MLE we observe in Figure \ref{fig:basic-sm} that the NC coefficients are overestimated.
We wondered if the fit was mainly impaired by trials with unrealistically high activity as in Figure \ref{fig:geometry}E. But that does not seem to be the case, because the fit remains low with MLE ($R^2=-0.78$) even when we discard trials where the firing probability of a neuron is higher than $0.85$ for $10$ consecutive time steps. 


\begin{table}
\centering
\caption{Performance summary on the test set when fitting RSNN models to the V1-dataset. The precise definition of the performance metrics are given in Appendix \ref{sec:app-performance-metrics}. The standard deviation across neurons is provided for the PSTH correlation. The variability across different network initialization is relatively low in comparison with the difference across algorithms, for instance we computed the standard deviation of $\LL_{MLE}$ over $3$ seeds for MLE and MLE+PSTH+NC${}_{MSE}$ and found respectively $2\cdot 10^{-5}$ and $3 \cdot 10^{-4}$. For the noise correlation $R^2$, the standard deviation was $9 \cdot 10^{-3}$ for MLE+PSTH+NC${}_{MSE}$.}
\label{tab:results_V1}
\begin{tabular}{c|ccc}
Method               & \begin{tabular}[c]{@{}c@{}}PSTH correlation\end{tabular} & \begin{tabular}[c]{@{}c@{}}Noise Correlation \\ ($R^2$)\end{tabular} & \begin{tabular}[c]{@{}c@{}} Negative log-likelihood\\
$\LL_{MLE}$ (on test set)\end{tabular}  \\ 
\hline
MLE  & $0.67 \pm 0.16$ & $-0.91$ & $0.370$  \\[0.5ex]
PSTH & $0.72 \pm 0.15$ & $-3.3$ & $0.44$  \\[0.5ex]
PSTH+NC${}_{MSE}$     & $0.69 \pm 0.15$ & $0.80$ & $0.50$ \\[0.5ex]
MLE+PSTH+NC${}_{MSE}$ & $0.69 \pm 0.15 $ & $0.73$ & $0.370$ 
\end{tabular}
\end{table}

We report in Figure \ref{fig:gan} the PSTH correlation and noise-correlation in a different format to enable a qualitative comparison with the results obtained with a spike-GAN on the same dataset (see Figure~3B,C from \cite{ramesh2019adversarial}).
The fit is qualitatively similar even if we do not need a separate discriminator network.
Also our RSNN model is better suited to make interpretations about the underlying circuitry because it models explicitly the neural dynamics and the recurrent interactions between the neurons (whereas a generic stochastic binary CNN without recurrent connections was used with the spike-GAN).

We also compare our approach with the 2-step method which is a contemporary alternative to MLE for fitting RSNNs \cite{mahuas2020new}.
The PSTH and noise correlation obtained with the 2-step method were measured on another dataset of 25 neurons recorded in-vitro in the retina of the Rat \cite{deny2017multiplexed}.
We trained our method on the same dataset under the two stimulus conditions and a quantitative comparison is summarized in Table \ref{tab:2step_result}. Under a moving bar stimulus condition we achieve a higher noise correlation ($3\%$ increase) and a higher PSTH correlation ($19$\% increase). But this difference might be explained by the use of a linear-simulus filter \cite{mahuas2020new} instead of a CNN.
Under a checkerboard stimulus condition, the 2-step method was tested with a CNN but we still achieve a better noise-correlation ($5 \%$ improvement) with a slightly worse PSTH correlation ($2\%$ decrease).
Another difference is that it is not clear how the 2-step method can be extended to model the activity of hidden neurons as done in the following section.

In summary, this section shows that using a differentiable simulator and simple sample-and-measure loss functions leads to a competitive generative model of neural activity.
The approach can also be generalized to fit single-trial statistics as explained in the Appendix \ref{sec:suppl-sample-and-measure} and Figure \ref{fig:mutli-step}. 


\section{Model identification}

Beyond simulating realistic activity statistics,
we want the RSNN parameters to reflect a truthful anatomical circuitry or realistic neural properties.
To test this, we consider a synthetic dataset generated by a \emph{target network} for which we know all the parameters.
We build this target network by fitting it to the V1-dataset and sample from this model a synthetic dataset of similar size as the V1-dataset ($80$ training trials of approximately $30$ seconds). Since our target network can generate as much data as we want, we simulate a larger test set of $480$ trials and a larger validation set of $40$ trials. 
We then fit \emph{student networks} on this synthetic dataset and compare the parameters $\phi$ of the \emph{student} networks with the ground-truth parameters $\phi^*$ of the \emph{target} networks.

\begin{figure}
  \centering
  \includegraphics[width=\textwidth]{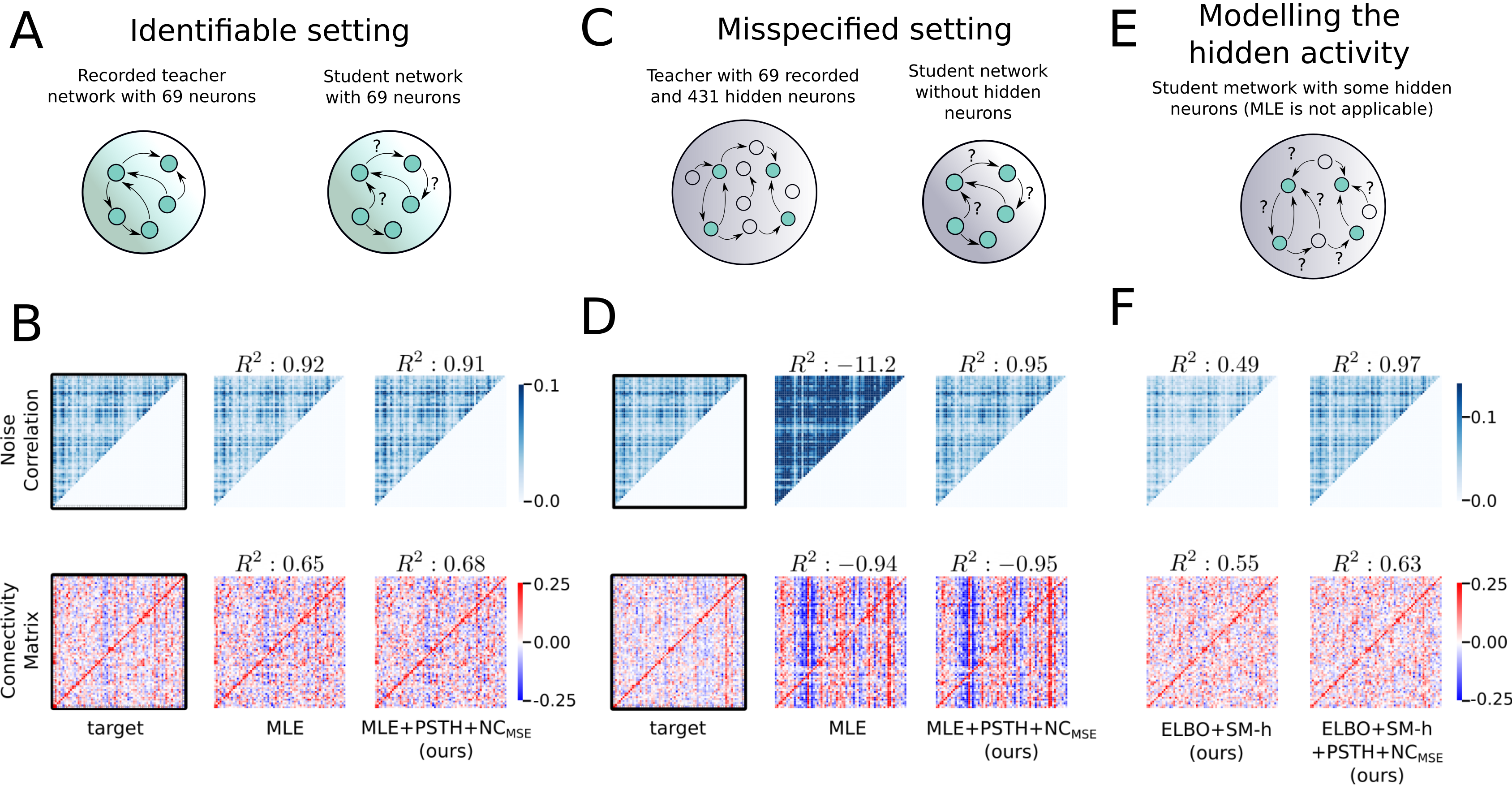}
  \caption{\textbf{A, B}) Results of the model identification experiment in the fully identifiable setting. \textbf{A} summarizes the simulation setup. The results are shown in \textbf{B}: in the first row, we show the noise correlation matrix. In the second row, we show the connectivity matrix $\sum_{d} W_{ji}^d$ where the x- axis indicates pre-synaptic neuron $i$ and y-axis indicates post-synaptic neuron $j$.
  \textbf{C, D}) Same as \textbf{A, B} but in a misspecified setting: the target network has $500$ neurons and student networks have $69$ neurons. 
  \textbf{E, F}) Same as before with the same target network as in \textbf{C, D}, but the student has some hidden neurons.
  When averaging $R^2$ of the connectivity matrices over 5 different student network initialization,
  we find $R^2=0.56\pm0.0069$ for ELBO+SM-h and $0.63\pm0.013$ for ELBO+SM-h+PSTH+NC$_{MSE}$.
  For the noise correlation, we find $0.66\pm0.1$ for ELBO+SM-h and $0.95\pm0.2$ for ELBO+SM-h+PSTH+NC$_{MSE}$.
  Here the number of hidden neurons in the student network is the same as in the target network, but this is not necessary as seen in Table \ref{tab:hidden_neurons}.
  The connectivity matrices are displayed with higher resolution in Figure \ref{fig:high-resolution} and \ref{fig:high-resolution-easy}.
  }
  \label{fig:model-identification}
\end{figure}

\paragraph{Well specified model without hidden neurons}
As a first comparison we consider the simplest case where the target network is fully observed: the target network consist of $69$ visible neurons and each student network is of the same size.
This is in fact the ideal setting where the log-likelihood is tractable and the MLE enjoys strong theoretical guarantees.
In particular if the CNN weights are not trained and are copied from the target-network, the loss function $\LL_{MLE}$ is convex with respect to the remaining RSNN parameters $\phi$ and the target network is identifiable \cite{paninski2004maximum}. 
The resulting fitting performance is summarized in Figure \ref{fig:model-identification}A where we show the NC matrix and the connectivity matrix $(\sum_d W_{i,j}^d)$ for the target network and two students networks. We do not show the PSTH because all methods already fit it well on the V1-dataset (see Table \ref{tab:results_V1}).
In this setting, combining $\LL_{NC}$ and $\LL_{PSTH}$ with $\LL_{MLE}$ brings almost no advantage: the MLE already provides a good reconstruction of the NC and connectivity matrices.

\paragraph{Model misspecification when ignoring hidden neurons}
From these results we hypothesize that this fully identifiable setting does not capture the failure of MLE observed with real data because the recorded neurons are embedded in a much larger biological network that we cannot record from.
To model this, we construct another synthetic dataset based on a larger target network of $500$ neurons where the first $69$ neurons are fitted to the neurons recorded in the V1-dataset and the remaining $431$ are only regularized to produce a realistic mean firing rate (see appendix for simulation details).
As in the standard setting where one ignores the presence of hidden neurons, we first consider that the student networks model only the first $69$ visible neurons. This model is therefore misspecified because the number of neurons are different in the target and student networks, hence this setting is well described by the scenario sketched in Figure \ref{fig:geometry}B.

The results are shown in Figure \ref{fig:model-identification}B. We found that MLE is much worse than the sample-and-measure method in this misspecified setting and the results resemble better what has been observed with real data. With MLE the noise-correlation coefficient are over estimated and the overall fit is rather poor (negative $R^2$), but it significantly improves after adding the sample-and-measure loss functions ($R^2=0.95$). This suggest that ignoring the impact of hidden neurons can explain the failure of MLE experienced in the real V1-dataset.
We find little relationship between the student and teacher connectivity matrices (only the connectivity between visible neurons are compared, see Figure \ref{fig:basic-sm}). This suggests that the standard strategy, where the hidden neurons are ignored, is unlikely to be informative about true cortical connectivity.

\paragraph{Well specified model with hidden neurons}
To investigate whether including hidden neurons leads to more truthful network models, we take the same target network of $500$ neurons and fit now student networks of the same size ($500$ neurons) but where only the first $69$ are considered visible (Figure \ref{fig:model-identification}C).
Since the model is well specified but data about the hidden neurons is missing, this experiment is well summarized by the scenario of Figure \ref{fig:geometry}C.
We use $\LL_{ELBO}$ for the visible units and we add a sample-and-measure loss function $\LL_{SM-h}$ to constrain the average firing rate of the hidden neurons which are completely unconstrained otherwise (see appendix). As seen in Figure \ref{fig:model-identification}C, it yields more accurate NC matrix ($R^2=0.49$) and connectivity matrix ($R^2=0.55$) compared to the previous misspecified models which did not include the hidden neurons.
When we add sample-and-measure loss functions to fit the PSTH and NC of the visible neurons, the noise-correlation matrix and connectivity matrix are fitted even better ($R^2=0.97$ and $R^2=0.63$).
Quantitatively, the $R^2$ for the connectivity matrices are almost as high as in the easy setting of panel A where the target network is fully-visible and identifiable.
Although the student network had exactly the same number of hidden neurons in Figure \ref{fig:model-identification} E-F, the success is not dependent on the exact number of hidden neurons as shown in Table \ref{tab:hidden_neurons}. Rather, assuming a small hidden population size of only 10 neurons was enough to alleviate the failure observed in the absence of hidden neurons in Figure \ref{fig:model-identification} C-D. Table \ref{tab:hidden_neurons} also shows however that the accuracy of the reconstruction improves substantially if the hidden population is made larger in the student network.

\section{Discussion}

We have introduced the sample-and-measure method for fitting an RSNN to spike train recordings. This method leverages deep learning software and back-propagation for stochastic RSNNs to minimize sample-and-measure loss functions.
A decisive feature of this method is to model simply and efficiently the activity of hidden neurons. We have shown that this is important to reconstruct trustworthy connectivity matrices in cortical areas.
We believe that our approach paves the way towards better models with neuroscientifically informed biases to reproduce accurately the recorded activity and functional connectivity.
Although we have focused here on GLMs, PSTH and NC, the method is applicable to other spiking neuron models and statistics.

\paragraph{Perspective} 
One of the promising aspects of our method is to fit models which are much larger.
One way to do this, is to combine neurons from separate sessions in a single larger network by considering them alternatively visible or hidden.
This problem was tackled partially in \cite{sorochynskyi2021predicting,turag2014inferring}.
It is natural to implement this with our method and we believe that it is a decisive step to produce models with a dense coverage of the recorded areas.

To investigate if our method is viable in this regime we carried out a prospective scaling experiment on a dataset from the Mouse brain recorded with multiple Neuropixels probes across 58 sessions \cite{siegle2021survey}. The goal of this scaling experiment is only to evaluate the amount of computing resources required to fit large networks. We ran three fitting experiments with $2$, $10$ and $20$ sessions respectively. Focusing on neurons from the visual cortices, it yielded models with $527$, $2219$ and $4995$ neurons respectively. Each simulation could be run on a single NVIDIA V100 GPU and running $100$ training epochs took approximately $4$, $12$ and $36$ hours respectively.
We conclude that this large simulation paradigm is approachable with methods like ours and we leave the fine-tuning of these experiments and the analysis of the results for future work.


\begin{ack}
This research was supported by Swiss National Science Foundation (no. 200020\_184615) and the Intel Neuromorphic Research Lab. Many thanks to Christos Sourmpis, Gabriel Mahuas, Ulisse Ferrari, Franz Scherr and Wolfgang Maass for helpful discussions.
Special thanks to Stéphane Deny, Olivier Marre and Ulisse Ferrari for sharing with us the Retina dataset and to Matthew Smith and Adam Kohn for making their dataset publicly available.

\paragraph{Authors contributions} GB and SW conceived the project initially. SW did most of the simulations under the supervision of GB. All authors contributed significantly to the theory and the writing.
\end{ack}

\small
\bibliographystyle{unsrt}
\bibliography{library}

\newpage

\appendix

\newcommand{\toptitlebar}{
  \hrule height 4pt
  \vskip 0.25in
  \vskip -\parskip%
}
\newcommand{\bottomtitlebar}{
  \vskip 0.29in
  \vskip -\parskip
  \hrule height 1pt
  \vskip 0.09in%
}

\newpage
\setcounter{page}{1}
\appendix
\begin{center}
{\Large \textbf{Appendices of:\\}}
\vspace{0.5cm}

\toptitlebar
{\LARGE \textbf{\newtitle}}
\bottomtitlebar
\vspace{0.5cm}
\end{center}

\section{Datasets}
\label{sec:app-datasets}

\paragraph{V1-dataset}
The dataset we used was collected by Smith and Kohn \cite{kohn2016utah} and is publicly available at: http://crcns.org/data-sets/vc/pvc-11. In summary,  macaque monkeys were anesthetized  with Utah arrays placed in the primary visual cortex (V1). In our analysis, we considered population spiking activity of monkey-I in response to a gray-scale natural movie. The movie is about a monkey wading through water.
It lasts for 30 seconds (with sampling rate 25Hz) and was played repeatedly for 120 times. Similarly as in \cite{ramesh2019adversarial}, we used the last 26 seconds of the movies and recordings.
Each frame of the movie has $320 \times 320$ pixels and we downsampled them to $27 \times 27$ pixels.
We used the recording from the 69 neurons with time bins $40$ms and considered that there cannot be more than one spike per bin ($5\%$ of the time bins had more than one spike).

\paragraph{Synthetic dataset}
Two target networks are trained using the V1-dataset: one with no hidden neuron and one with 431 hidden neurons which makes 500 neurons in total. To build the target network without hidden neurons, we fitted a network with the loss function $\LL_{MLE+PSTH+NC}$. For the target network with hidden neurons, we train a network using $\LL_{MLE+SM-h+PSTH+NC}$.

\paragraph{Retina dataset}
The data we used were the same as \cite{mahuas2020new} and was initially published in \cite{deny2017multiplexed}. It was generously shared with us privately.
It contained recorded spike trains for
25 OFF Alpha retinal ganglion cells' in the form of binarized spike counts in 1.667ms bins.
There were two stimulus conditions.
For the checkerboard, the unrepeated movie (1080s) plus one repeated movie (600s in total for 120 repetitions) were used for training and the other repeated movie (480s in total for 120 repetitions) were used for testing. For the moving bar, the unrepeated movie (1800s) plus one repeated movie (166s in total for 50 repetitions) were used for training and the other repeated movie (322s in total for 50 repetitions) were used for testing.


\section{Simulation details}
\label{sec:app-simulation-details}

\paragraph{For the V1- and synthetic datasets}
The model combines a spatio-temporal CNN and an RSNN.
Input to the CNN consists of 10 consecutive movie frames. The CNN has 2 hidden layers and its output is fed into the RSNN. To feed the images to the CNN the 10 gray-scaled images are concatenated on the channel dimension. The two hidden layers include convolution with 16 and 32 filters, size 7 by 7 (with padding) followed by a ReLU activation function and then a MaxPool layer with kernel size 3 and stride 2 as in \cite{kindel2019using}.
The weights from the CNN to the RSNN are initialized with a truncated normal distribution with standard deviation $\frac{1}{\sqrt{n_{in}}}$ where $n_{in}$ in the number of inputs in the weight matrix. The tensor of recurrent weights $\bm W$ consider spike history of last $9$ frames ($d_{\max}$) and the weight distribution is initialized as a truncated normal distribution with standard deviation $\frac{1}{\sqrt{d_{\max} n_{\HH + \VV}}}$. The bias $\bm b$ is initialized with zero. The voltage threshold $v_\mathrm{thr}$ is set to 0.4 and the dampening factor $\gamma$ is 0.3. We used an Adam optimizer. More hyper-parameters like learning rates are given in Table \ref{tab:hyperparam_V1} and Table \ref{tab:hyperparam_syn}.
To implement the loss $\LL_{MLE+PSTH+NC}$, we process the CNN once and simulate the RSNN twice. Once the RSNN is clamped to the recorded spikes to compute $\LL_{MLE}$ or $\LL_{ELBO}$, the second time the sample and generated "freely" to compute $\LL_{PSTH}$ and $\LL_{NC}$.

\paragraph{For the retina dataset experiment}\label{para:retina_exp}
Since the time step is much smaller for the Retina dataset than for the V1-dataset (1.67ms rather than 40ms) the temporal filters have to be larger to take into account the full temporal context.
For both the receptive fields of the CNN and the tensor $\bm W$ we chose to cover time scales that are consistent with \cite{mahuas2020new}.
Hence we adapted the model architecture from the previous paragraph and added as a first layer of the CNN a causal temporal convolution (Conv1D with appropriate padding). The temporal convolution has a receptive filed of 300 time bins and outputs 16 filters. In the RSNN we choose  $d_{\max}=24$ so that the spike history filter covers around $40$ms.
Two fitting algorithms were tested, one with $\LL_{MLE}$ and the other one with $\LL_{MLE+single-trial+NC}$. The loss function $\LL_{single-trial}$ is used to fit single-trial statistics as defined in Appendix \ref{sec:suppl-sample-and-measure} and we used it here to replace $\LL_{PSTH}$ because some movies of training dataset are unrepeated and we saw in Figure \ref{fig:mutli-step} that it fits the PSTH almost as well as $\LL_{PSTH}$. To implement the loss $\LL_{MLE+single-trial+NC}$, we process the CNN once and simulate the RSNN twice for $T$ time steps.
The first time the RSNN is clamped to the recorded spikes for $T_{gt}$ time steps and then clamping is terminated and the RSNN generates samples "freely" for the next $T-T_{gt}$ time steps. For the first $T_{gt}$ time steps, $\LL_{MLE}$ is computed. And for the rest $T-T_{gt}$ time steps where the activity is not clamped, $\LL_{single-trial}$ is computed as the cross entropy between $\bm z^\DD$ and the spike probabilities. 
We also run the RSNN a second time with the same CNN input and without any clamping to compute $\LL_{NC}$. 
For each gradient descent step, we sample uniformly from the dataset a batch of size $K^\DD= K_{m} \times K_{t}$ gathering truncated movie clips and corresponding spikes with $K_{m}$ different starting time points and from $K_{t}$ different movies. The hyper-parameters can be found in Table \ref{tab:hyperparam_2step}.

\section{Performance metrics}
\label{sec:app-performance-metrics}

For the definition of our performance metrics we use the following notations. The trial averaged firing probability of neuron $i$ in the time bin $t$ is denoted $\bar{z}_{t,i} = \frac{1}{K}\sum_{k} z_{t,i}^{k}$ where $z_{t,i}^{k} \in \{0,1\}$ is the spike and $K$ is the number of trials.
Neuron $i$'s mean firing rate is further computed as $\bar{z}_i=\frac{1}{T}\sum_t \bar{z}_{t,i}$ where $T$ is the number of time steps.

\paragraph{Peristimulus time histogram (PSTH) correlation}
The fit performance of the PSTH is measured by the Pearson's correlation between the simulated PSTH and the recorded PSTH. Hence for each neuron the PSTH correlation is defined by:
\begin{equation}
\rho_{i}^{\mathrm{PSTH}} = \frac{\sum_{t}\left(\bar{z}_{t,i}-\bar{z}_{i}\right)\left(\bar{z}^\mathcal{D}_{t,i}-\bar{z}^\mathcal{D}_{i}\right)}{\sqrt{\sum_{t}\left(\bar{z}_{t,i}-\bar{z}_{i}\right)^2}\sqrt{\sum_{t}\left(\bar{z}^\mathcal{D}_{t,i}-\bar{z}^\mathcal{D}_{i}\right)^2} }~,
\end{equation}
and a slightly better estimator of the asymptotical Pearson correlation which is less noisy can be estimated by replacing $\bar{z}_{t,i}$ and $\bar{z}_{i}$ with $\bar{\sigma}_{t,i}$ and $\bar{\sigma}_{i}$.

\paragraph{Noise-correlation matrix}
Pairwise noise correlations are computed as in \cite{sorochynskyi2021predicting}. We first define total covariance $M_{i,j}^{\mathrm{total}}$ and noise covariance $M_{i,j}^{\mathrm{noise}}$ between neuron $i$ and $j$.
\label{sec:measurement}
\begin{eqnarray}
M_{i,j}^{\mathrm{total}} & = & \frac{1}{T K}\sum_{t,k} \left(z_{t,i}^{k}-\bar{z}_{i}\right)\left(z_{t,j}^{k}-\bar{z}_{j}\right) \\
M_{i,j}^{\mathrm{noise}} & = & \frac{1}{T K}\sum_{t,k} \left(z_{t,i}^{k}-\bar{z}_{t,i}\right)\left(z_{t,j}^{k}-\bar{z}_{t,j}\right)
\label{eq:m_ij_noise}
\end{eqnarray}
Then in the performance tables we report the normalized noise correlation $\mathcal{M}_{i,j}^{\mathrm{noise}}$ for $i\neq j$:
\begin{equation}
\mathcal{M}_{i,j}^{\mathrm{noise}} = \frac{M_{i,j}^{\mathrm{noise}}}{\sqrt{M_{i,i}^{\mathrm{total}} M_{j,j}^{\mathrm{total}}}}~.
\label{eq:nc-normalized}
\end{equation}
We then define the coefficient of determination of the NC matrix $R^2$ as in \cite{mahuas2020new}.
Given the NC matrices computed from the data $\mathcal{M}_{i,j}^{\mathrm{noise},\DD}$
and the NC matrix obtained from the simulation $\mathcal{M}_{i,j}^{\mathrm{noise},\phi}$ 
we define $\mathcal{M}^{\mathrm{noise},\DD}=\frac{1}{{n_\VV}^2} \sum_{i,j} \mathcal{M}_{i,j}^{\mathrm{noise},\DD}$ and:
\begin{equation}
    R^2 = 1 - \frac{\sum_{i,j}\left( \mathcal{M}_{i,j}^{\mathrm{noise},\DD} - \mathcal{M}_{i,j}^{\mathrm{noise},\phi} \right)^2 }{
    \sum_{i,j} \left(\mathcal{M}_{i,j}^{\mathrm{noise},\DD}- \mathcal{M}^{\mathrm{noise},\DD}
    \right)^2}~.
\end{equation}

\section{Derivations of the loss functions}
\label{sec:suppl-sample-and-measure}

\paragraph{Normalization of the sample-and-measure functions} Most sample-and-measure may be defined one multiplicative constant away from their formal definition.
For instance when computing $\LL_{MLE}$ we compute the binary cross entropy between the relevant probabilities aggregate them by taking the mean and not the sum. We find the resulting number to be easier to interpret because is it independent from the number of trials and the number of time steps.

\paragraph{Noise correlation}
We tested two sample-and-measure loss function for the noise correlation. We explain here why the Monte-Carlo estimate of the simulated statistics is unbiased for $\LL_{NC}$ but the same argument applies to $\LL_{NC_{MSE}}$.

We consider the statistics $\TT(\bm z)_{ij} = \frac{1}{K T} \sum_{t,k} z_{t,i}^{k} z_{t,j}^{k}$ which measure the frequency of coincident spikes between neurons $i$ and $j$.
Since $z_{t,i}^{k}$ and $z_{t,j}^{k}$ are independent given the past, we have $\EE_{\PP_{\phi}} \left[ z_{t,i}^{k} z_{t,j}^{k}  \right]$ = $\EE_{\PP_{\phi}}\left[ \sigma(u_{t,i}^{k}) \sigma(u_{t,j}^{k}) \right]$ so we use the following Monte-Carlo estimate $\pi_{i,j}^\phi = \frac{1}{K T} \sum_{t,k} \sigma(u_{t,i}^{k}) \sigma(u_{t,j}^{k})$ to evaluate the expected simulated statistics in equation \eqref{eq:sm-with-stats}.
Choosing the dissimilarity $d$ to be the cross entropy and denoting $\pi_{i,j}^\DD = \TT(\bm z^\DD)_{ij}$ we define:
\begin{equation}
\mathcal{L}_{NC} = \sum_{i,j} CE(\pi_{i,j}^\DD, \pi_{i,j}^\phi)
\end{equation}
As an attempt to replace the terms in $\LL_{NC_{MSE}}$ which take into account the correlation from the PSTH, we tried to add a related correction term in $\LL_{NC}$.
To do do we considerd another loss $\mathcal{L}_{NC \mhyphen \mathrm{shuffled}}$ which is computed like $\LL_{NC}$ but where we shuffle the trial identities in $\bm z_i$ and not in $\bm z_j$. It seems that it was not as efficient as $\LL_{NC_{MSE}}$.


\paragraph{Single-trial statistics}
Since both PSTH and NC are trial-averaged statistics we wondered whether another simple measuring model could account for single-trial statistics.
We therefore considered the following problem which is notoriously challenging for the MLE~\cite{hocker2017multistep}: we clamp the network to the recorded data until time $t$ and generate a simulated spike train for $t'>t$. With MLE the network activity quickly diverges away from the real data.
To measure this quantitatively we estimate the multi-step log-likelihood $\mathcal{P}_\phi(\bm z_{t+\Delta t}^\DD | \bm z_0^\DD \cdots \bm z_t^\DD)$. It is intractable but an unbiased Monte-Carlo estimate can be computed. The multi-step log-likelihood drops quickly as $\Delta t$ increases as expected for MLE in Figure \ref{fig:mutli-step}.

To resolve that issue, we first suggest an extension of the definition of $\LL_{SM}$ in equation \ref{eq:sm-with-stats} which formalizes the clamping condition:
\begin{equation}
    \LL_{SM} = d \big(~ \EE_{\PP_\DD} \left[~ \TT(\bm z) ~|~ \bm c ~\right],~  \EE_{\PP_\phi} \left[~ \TT(\bm z) ~|~ \bm c~\right] ~\big)~,
    \label{eq:sm-with-condition}
\end{equation}
where we have introduced a condition $\bm c$ into the expectations. All the theory and the geometrical interpretations can be extended with this conditioning, but this allows to formalize that the visible units can be clamped to the recorded data. 
For instance if we choose $\bm c$ such that $\bm z^\VV_{1:t} = \bm z^{\DD}_{1:t}$ we formalize a sample-and-measure loss function for which the visible units are clamped until time $t$.

Back to the problem of fitting the multi-step log-likelihood, we consider the sample-and-measure loss function where $\TT$ is identity, $\sigma(\bm u)$ is the Monte-Carlo estimator and $d$ is the cross-entropy. It yields:
\begin{equation}
    \LL_{single \mhyphen trial} =
    CE(\bm z^\DD, \sigma(\bm u^\VV))~,
\end{equation}
which is pretty much computed like $\LL_{MLE}$ but where the data is only clamped until time $t$.
Note that since the statistics $\TT$ do not involve a trial average, the computation of the expectation is not very precise but it may be improved for the expectation $\EE_{\PP_\phi}$ by averaging over multiple batches clamped to the same data. Although this is an interesting direction we did not try it and always sample a single batch per clamping condition.
When using this loss function, we see in Figure \ref{fig:mutli-step}B that MLE only better just at the first time step after the clamping terminates and optimizing $\LL_{single \mhyphen trial}$ makes better prediction after that.
To provide a meaningful baseline we show the m-step likelihood obtained with a theoretical model fitting perfectly the PSTH without being aware of the clamping history. The multi-step likelihood obtained with $\LL_{single \mhyphen trial}$ is above this baseline for $5$ time-steps ($200$ms) on the training set proving that the model tries to make a clever usage of the trial specific firing history up to this duration.

\paragraph{Derivation of the ELBO}
Like for capturing single trial statistics, the most natural way to fit neural activity in the presence of hidden neurons is to minimize the cross-entropy between the visible spikes and their probability while sampling from the hidden neurons. Here we want to show that this is actually the negative of a variational lower bound of the maximum likelihood.
Following \cite{jordan1999introduction}, for any distribution $q(\bm z^{\HH})$ of the hidden neural activity we have:
\begin{eqnarray}
\log \PP_\phi (\bm z^\DD)
& = & 
\log \sum_{\bm z^\HH} \PP_\phi (\bm z^\DD, \bm z^\HH) \\
& = & 
\log \sum_{\bm z^\HH} q(\bm z^\HH) \frac{\PP_\phi (\bm z^\DD, \bm z^\HH)}{q(\bm z^\HH)}
\\
& \geq &
\sum_{\bm z^\HH} q(\bm z^\HH)
\log  \frac{\PP_\phi (\bm z^\DD, \bm z^\HH)}{q(\bm z^\HH)}
\end{eqnarray}
Writing $\bm z_t$ as the concatenation of $\bm z^\DD_t$ and $\bm z^\HH_t$, we now choose specifically $q$ so that for all $t$: $q(\bm z^{\HH}_t)=\PP_\phi(\bm z^\HH_t | \bm z_{1:t-1})$, using the factorization and seeing that the probability factorizes as follows: $\PP_\phi(\bm z^\DD, \bm z^\HH) = \prod_t \PP_\phi(\bm z^\DD_t, \bm z^\HH_t | \bm z_{1:t-1}) = \prod_t
\PP_\phi(\bm z^\DD_t | \bm z_{1:t-1}) \cdot \prod_t
\PP_\phi(\bm z^\HH_t | \bm z_{1:t-1})$, some products inside the $\log$ are cancelling out and we found the lower bound:
\begin{eqnarray}
\log \PP_\phi (\bm z^\DD)
& \geq & \EE_{q}
\left[ \sum_t \log \PP_\phi (\bm z^\DD_t |\bm z_{1:t-1}) \right]\\
& = & -\EE_{q}
\left[ CE(\bm z^\DD, \sigma(\bm u^\DD)) \right],
\label{eq:eblo-derivation}
\end{eqnarray}
Interestingly, a similar loss function can also be formulated as a sample-and-measure loss function. To do so we consider the definition from equation \eqref{eq:sm-with-condition} with the condition $\bm c$ being $\bm z^\VV = \bm z^\DD$ meaning that all the visible units are clamped to the data. Choosing otherwise $\TT$ to be the identity and $d$ as the cross-entropy, we obtain the following loss function denoted as $\LL_{ELBO-SM}$:
\begin{equation}
\LL_{ELBO-SM}
=CE\left(\bm z^\DD, \EE_{\PP_\phi} \left[ \sigma(\bm u^\DD) \right] \right).
\label{eq:elbo-from-sm}
\end{equation}
Comparing the two loss functions we see that the essential difference is the placement of the expectation $\EE_{\PP_\phi}$.
In practice our current optimization minimizes \eqref{eq:eblo-derivation} rather than \eqref{eq:elbo-from-sm} because we sample a single trial for each clamping condition and apply stochastic gradient descent with momentum.
This implements implicitly the averaging of the gradients which corresponds better to the expectation from equation \eqref{eq:eblo-derivation}.
However it is also possible to minimize \eqref{eq:elbo-from-sm} by averaging the Monte-Carlo estimates obtained with multiple simulations with the same clamping condition. With enough sample it may provides a better estimate of the expectation $\EE_{\PP_\phi} \left[ \sigma(u^\DD) \right]$. 
The down side of this alternative is that it requires to sample more RSNN trajectories for each gradient update which may consume compute time inefficiently.
On the other hand, this might be relevant in another setting or at the end of training to benefit from the theoretical properties of the sample-and-measure loss function. We leave this to future work.

\paragraph{Regularization of the firing rate of hidden neurons}
When simulating hidden neurons which are never recorded it is desirable to insert that as much prior knowledge as possible about the hidden activity to keep the network model in a realistic regime.
The most basic prior is to assume that every neuron $i$ should have a realistic average firing rate, to implement this we design again a sample-and-measure objective as a variant of $\mathcal{L}_{PSTH}$.
This time we consider that the statistics $\TT$ are the average firing rate of a neuron $\TT(\bm z_i) = \sum_{t,k} z_{t,i}^{k}$.
This results in the objective $\mathcal{L}_{SM \mhyphen h}$ which is otherwise similar to $\mathcal{L}_{PSTH}$ as defined in equation  \eqref{eq:obj-psth}. Unfortunately the objective cannot be implemented as such because of one missing element: the empirical probability $\pi_i^\DD$ of a hidden neuron. Instead we simply take another neuron $j$ at random in the visible population and use this average firing rate in place of the probability $\pi_{i}^\DD$. In this way, the distribution of average firing rates across neurons of the hidden neurons is realistic at a population level because it becomes the same in the recorded population and in simulated population.

\renewcommand{\thesection}{S\arabic{section}}  
\renewcommand{\thetable}{S\arabic{table}}  
\renewcommand{\thefigure}{S\arabic{figure}}

\begin{table}
\centering
\begin{tabular}{c|ccccc}
Method      & \begin{tabular}[c]{@{}c@{}}learning \\ rate\end{tabular} & \begin{tabular}[c]{@{}c@{}}batch \\ size\end{tabular} & $\mu_{PSTH}$ & $\mu_{NC}$ & $\mu_{MLE}$  \\ [0.5ex]
\hline
MLE         & \multirow{5}{*}{1e-3}    & \multirow{5}{*}{20}  & 0            & 0          & 1            \\[1ex]
PSTH        &                                                       &                                                      & 1            & 0          & 0            \\[1ex]
MLE+PSTH        &                                                       &                                                      & 0.5            & 0.5          & 0            \\[1ex]
PSTH+NC     &                                                     &                                                   & 0.11          & 0.89          & 0            \\[1ex]
MLE+PSTH+NC &                                                      &                                                   & 0.1          & 0.5        & 0.4 \\[1ex]
MLE+PSTH+NC${}_{MSE}$ &                                                      &                                                   & 0.1          & 50        & 0.4 \\[1ex]
\end{tabular}
\caption{Hyper-parameter table used when fitting the V1-dataset (Figure \ref{fig:basic-sm}).}
\label{tab:hyperparam_V1}
\end{table}

\begin{table}
\begin{tabular}{c|cccccc}
Method           & learning rate           & batch size          & $\mu_{PSTH}$ & $\mu_{NC}$ & $\mu_{MLE}$ & $\mu_{SM-h}$ \\ [0.5ex] \hline
MLE              & \multirow{4}{*}{1.5e-3} & \multirow{4}{*}{20} & 0            & 0          & 1           & 0            \\[1ex]
MLE+PSTH+NC      &                         &                     & 0.1          & 0.7        & 0.2         & 0            \\[1ex]
MLE+SM-h         &                         &                     & 0            & 0          & 0           & 1e-3         \\[1ex]
MLE+SM-h+PSTH+NC &                         &                     & 0.1          & 0.7        & 0.2         & 1e-3 \\[1ex]       
\end{tabular}
\caption{Hyper-parameter table used when fitting the synthetic dataset (Figure \ref{fig:model-identification}). Early-stopping was used on 40 validation trials to prevent over-fitting.}
\label{tab:hyperparam_syn}
\end{table}

\begin{table}
\centering
\begin{tabular}{c|cc}
Number of Hidden Neurons               & \begin{tabular}[c]{@{}c@{}}Noise Correlation \\ $R^2$\end{tabular} & \begin{tabular}[c]{@{}c@{}} Connectivity Matrix \\
$R^2$\end{tabular}  \\ 
\hline
0  & $0.95$ & $-0.95$  \\[0.5ex]
10 & $0.96$ & $0.50$  \\[0.5ex]
200& $0.97$ & $0.59$ \\[0.5ex]
400& $0.95$ & $0.62$ 
\end{tabular}
\caption{Performance summary on the test set when fitting RSNN models with variable number of hidden neurons to the synthetic dataset. ELBO+SM-h+PSTH+NC${}_{MSE}$ method is used when hidden neurons are included, and MLE+PSTH+NC${}_{MSE}$ method is used when no hidden neurons are included.}
\label{tab:hidden_neurons}
\end{table}

\begin{table}
\centering
\begin{tabular}{c|c|ccccc}
Dataset               & Method &
$T_{gt}$  & $T$ & $K_m$  & $K_t$ & learning rate \\ [0.5ex] \hline
\multirow{2}{*}{Moving bars stimulus}  & MLE + single-trial + NC           & 45 & 50  & 20 & 8 & 5e-3          \\[1ex]
                                       & MLE                     & 50 & 50  & 20 & 8 & 5e-3          \\ [1ex]\hline
\multirow{2}{*}{Checkerboard stimulus} & MLE + single-trial + NC           & 75 & 80  & 30 & 1 & 1e-3         \\[1ex]
                                       & MLE                     & 75 & 75  & 8  & 4 & 1e-3          \\ [1ex]
\end{tabular}
  \caption{Hyper-parameter table for fitting the Retina Dataset, the definition of the hyper-parameter is given in Appendix \ref{sec:app-simulation-details}. The results are reported in Table \ref{tab:2step_result}.
  }
  \label{tab:hyperparam_2step}
\end{table}

\renewcommand{\figurename}{Figure} 
\setcounter{figure}{0}    

\begin{figure}
  \centering
  \includegraphics[width=\textwidth]{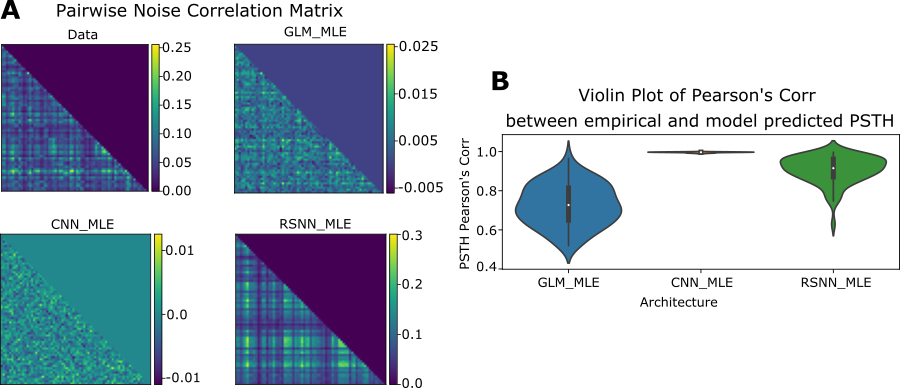}
  \caption{\textbf{Comparison of the network architecture}
  As a preliminary experiment we compared our network architecture against the official GLM code \cite{pillow2008spatio}. In our architecture a CNN replaces the spatio-temporal stimulus filter of the CNN, both models are fitted with MLE. \textbf{A} Noise correlation matrix of the different models. We included a control architecture where we pruned out the recurrent connection. It is called CNN because only the CNN parameters become relevant. \textbf{B} The PSTH correlation computed on the training set. The violon plot represents the distribution of neurons.}
  \label{fig:cnn}
\end{figure}

\begin{figure}
  \centering
  \includegraphics[width=\textwidth]{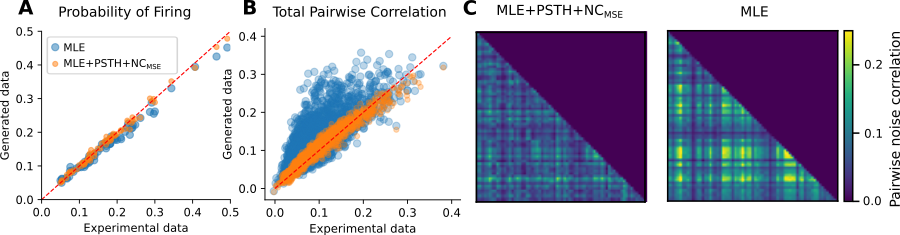}
  \caption{\textbf{Comparison with SpikeGAN} This figure is meant to be compared with the Figure 3 from \cite{ramesh2019adversarial}. In this other paper, the authors fitted a spike-GAN to the same dataset.
  We argue that the PTSH correlation and NC coefficients are as good qualitatively as the results obtained in \cite{ramesh2019adversarial}.
  }
  \label{fig:gan}
\end{figure}

\begin{figure}
  \centering
  \includegraphics[width=\textwidth]{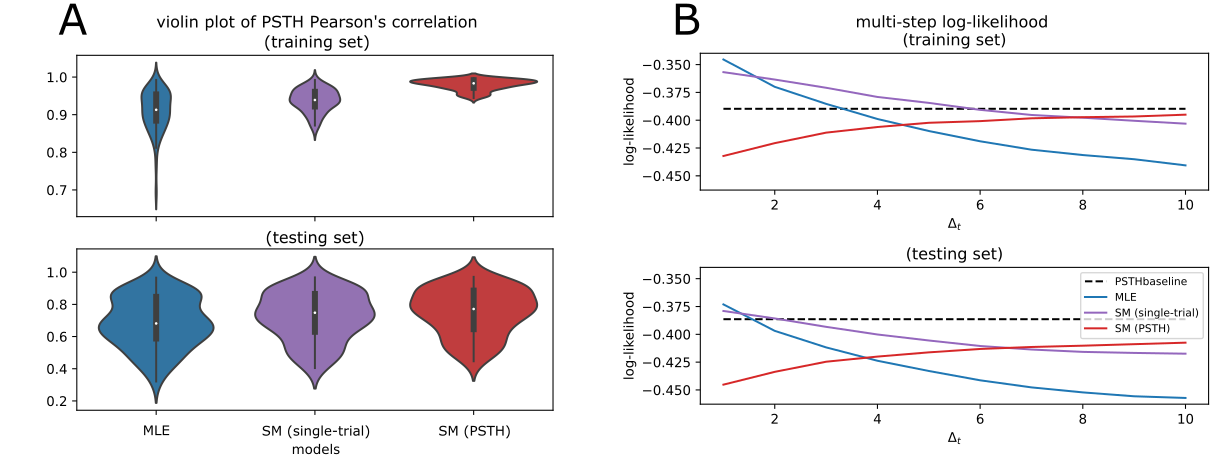}
  \caption{\textbf{Improving the mutli-step log-likelihood}
  We tackle the challenge identified in \cite{hocker2017multistep}.
  We evaluated the multi-step log-likelihood $\log \PP_\phi(\bm z^\DD_{t+\Delta t} | \bm z_0^\DD \cdots \bm z_t^\DD)$ as explained in the main text, and we trained two networks to minimize $\mathcal{L}_{MLE}$ and $\mathcal{L}_{single-trial}$ respectively. 
  \textbf{A}) The PSTH correlation of the different models trained in this context.
    \textbf{B}) The multi-step log-likelihood is reported for different models. The dashed baseline represent the ideal model which would always fire a spike with the true PSTH probability.
    The blue baseline is $\LL_{MLE}$ it has never seen self-generated activity during the training, so it's performance drops quickly when the network is not clamped anymore ($\Delta t >0$).
    The red-baseline is a model trained with $\LL_{PSTH}$ only, it is increasing because the model has never been clamped during training. Therefore it is not trained to be accurate right after the clamping terminates.
  }
  \label{fig:mutli-step}
\end{figure}

\begin{figure}
  \centering
  \includegraphics[width=\textwidth]{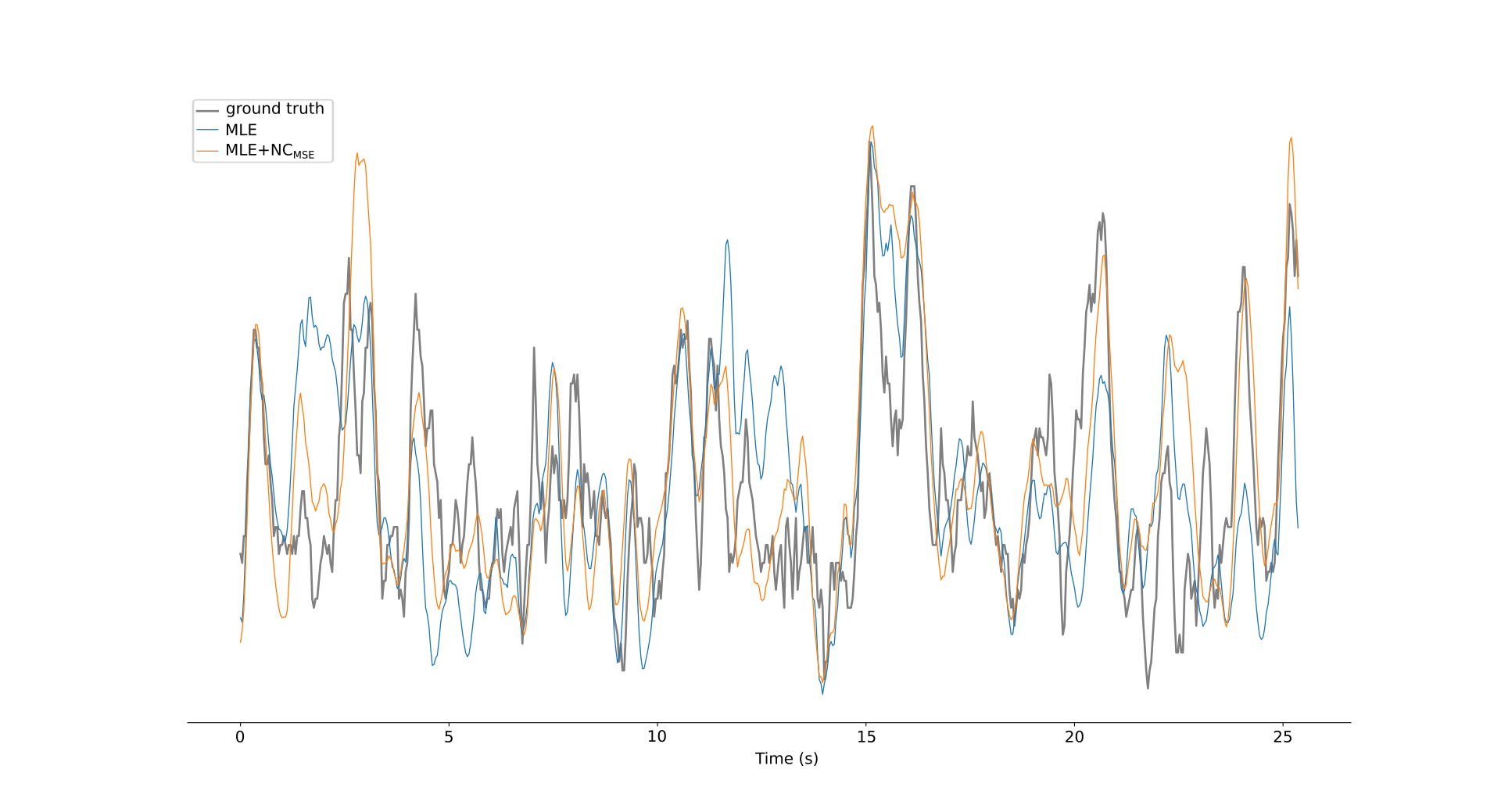} \caption{Example of PSTH obtained for a neuron from the V1-dataset. The PSTH are smoothed with a window size of 280ms.}

  \label{fig:psth-comparison}
\end{figure}

\begin{table}
\centering

\begin{tabular}{c|cc|cc}
Method        & \multicolumn{2}{c|}{Moving bars stimulus} & \multicolumn{2}{c}{Checkerboard stimulus}  \\
              & PSTH         & noise-corr.                & PSTH         & noise-corr.                 \\ [0.5ex] 
\hline
MLE + single-trial + NC & 0.91 ± 0.003 & 0.94                       & 0.85 ± 0.004 & 0.96                        \\[1ex] 
MLE           & 0.90 ± 0.002 & 0.91                       & 0.84 ± 0.003 & 0.96                        \\ [1ex] 
\hline
2-step (CNN) & -            & -                          & 0.87~± 0.04  & 0.91                        \\[1ex] 
2-step      & 0.72 ± 0.10  & 0.91                       & 0.81 ± 0.05  & 0.95                        \\[1ex] 
\end{tabular}
\caption{Performance comparison with the 2-step method \cite{mahuas2020new} on the Retina Dataset.
The performance of the 2-step dataset are taken from their paper. For this comparison we used their definition of $R^2$ which does not penalize a constant offset between prediction and target.
For historical reasons, we used here a sample-and-measure NC loss with cross-entropy and not mean-square error. It works but considering later simulations on other datasets we expect higher $R^2$ with mean-square error.}
\label{tab:2step_result}
\end{table}

\begin{figure}
  \centering
  \includegraphics[height=0.66\textheight]{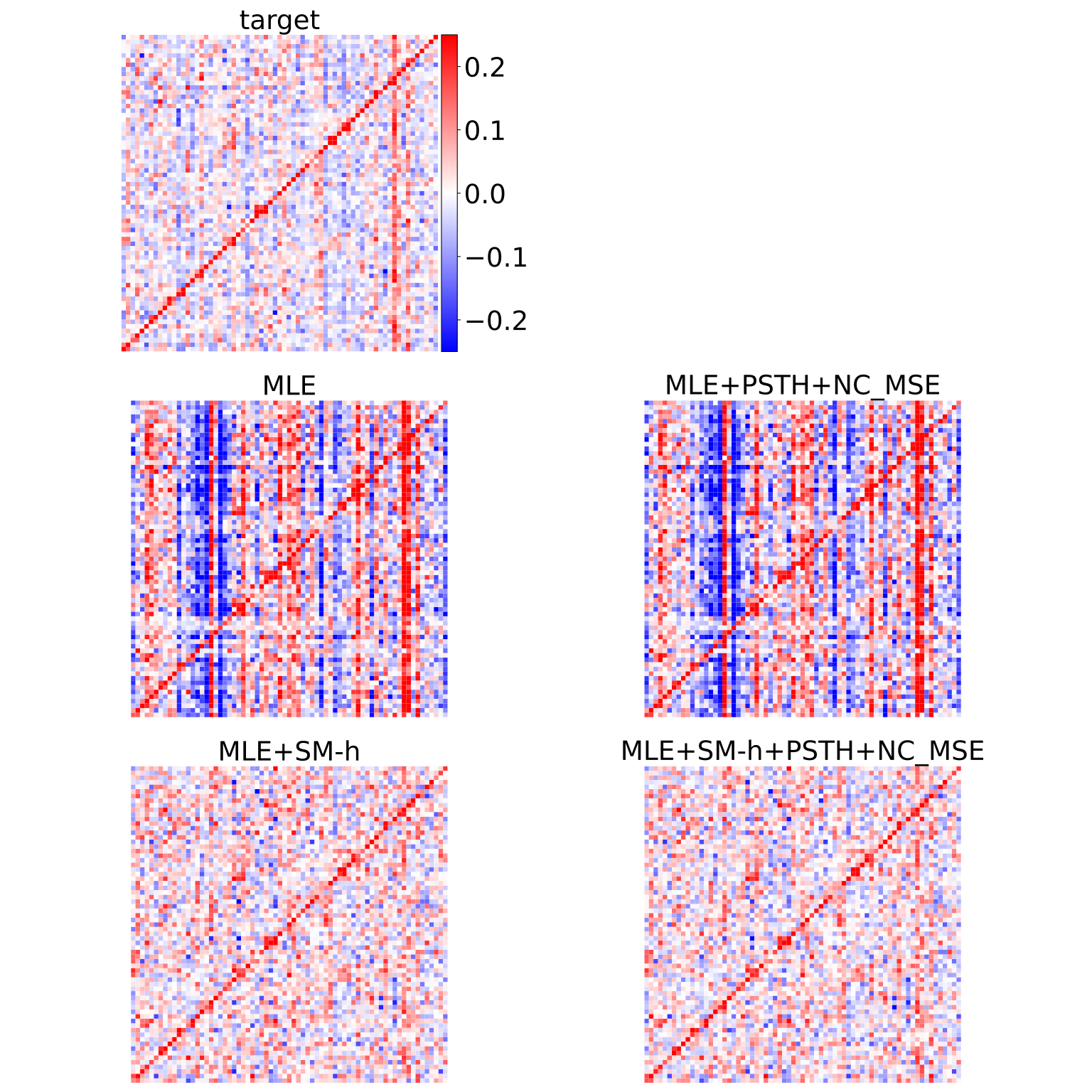}
  \caption{\textbf{High resolution plot of connectivity matrices in Figure \ref{fig:model-identification} (large target network)}
  }
  \label{fig:high-resolution}
\end{figure}

\begin{figure}
  \centering
  \includegraphics[height=0.44\textheight]{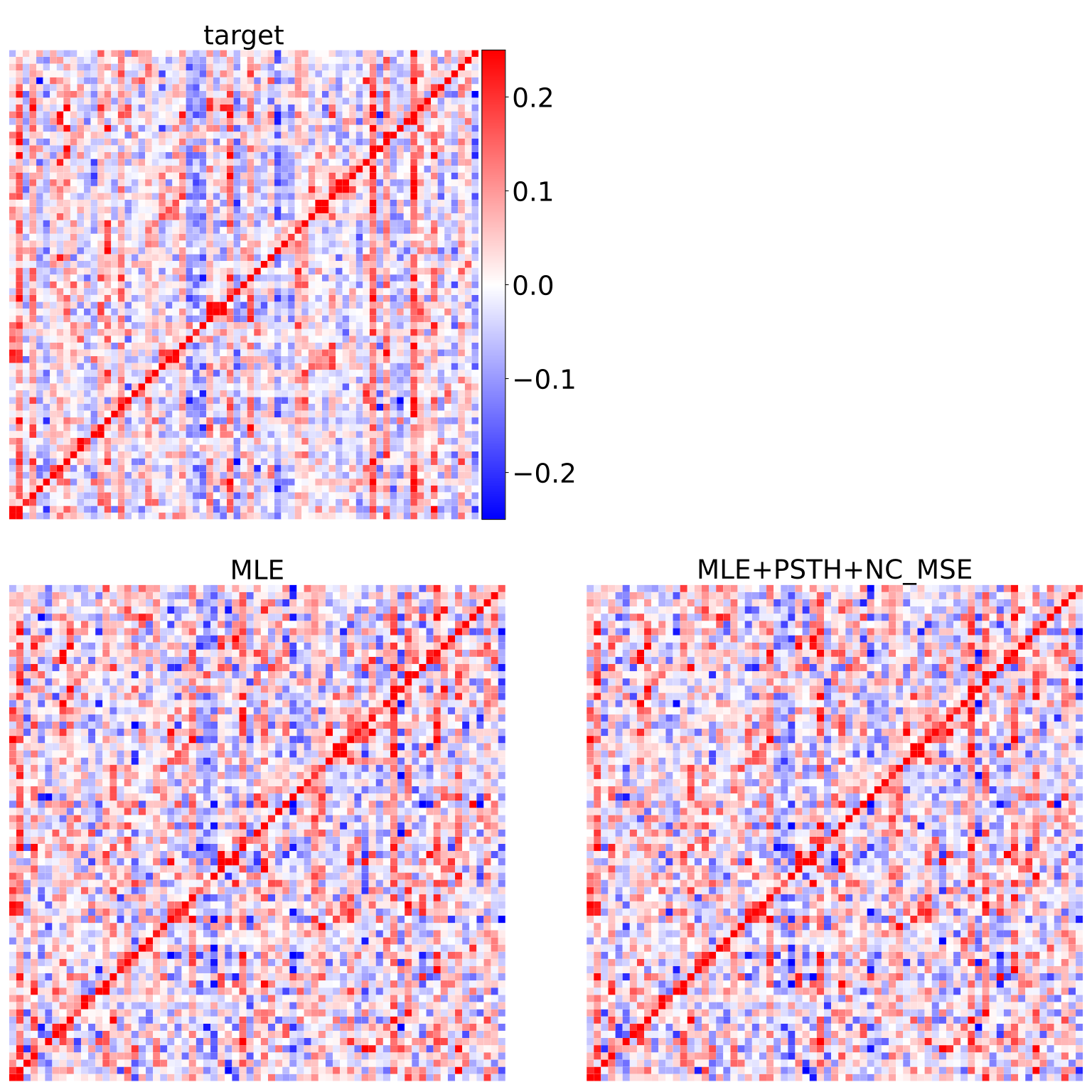}
  \caption{\textbf{High resolution plot of connectivity matrices in Figure \ref{fig:model-identification} (small target network)}
  }
  \label{fig:high-resolution-easy}
\end{figure}

\end{document}